# Spatial Priming Outperforms Semantic Prompting: A Grid-Based Approach to Improving LLM Accuracy on Chart Data Extraction

Andrei Lazarev
Department of Control and Applied Mathematics
Moscow Institute of Physics and Technology
Moscow, Russia
ORCID:0009-0008-1519-5820

Dmitrii Sedov
Moscow Institute of Physics and Technology
Moscow, Russia
ORCID:0009-0009-0964-7576

Alexander Galkin
Institute of Computer Science
Lipetsk State Technical University
Lipetsk, Russia
ORCID:0000-0002-4876-4865

***Abstract*—The automated extraction of data from scientific charts is a critical task for large-scale literature analysis. While multimodal Large Language Models (LLMs) show promise, their accuracy on non-standardized charts remains a challenge. This raises a key research question: what is the most effective strategy to improve model performance (high-level semantic priming) or low-level spatial priming? This paper presents a comparative investigation into these two distinct strategies. We describe our exploratory experiments with semantic methods, such as a two-stage metadata-first framework and Chain-of-Thought, which failed to produce a statistically significant improvement. In contrast, we present a simple but highly effective spatial priming method: overlaying a coordinate grid onto the chart image before analysis. Our quantitative experiment on a synthetic dataset demonstrates that this grid-based approach provides a statistically significant reduction in data extraction error (SMAPE reduced from 25.5% to 19.5%, $p < 0.05$) compared to a baseline. We conclude that for the current generation of multimodal models, providing explicit spatial context is a more effective and reliable strategy than high-level semantic guidance for this class of tasks.**

## I. Introduction

The modern scientific landscape is characterized by a sustained exponential growth of scholarly output, a phenomenon confirmed by large-scale bibliometric analyses of major citation indexes like Scopus [1]. Importantly, this quantitative growth in publications has been paralleled by an increasing reliance on data visualizations to convey complex findings. Scientific charts and graphs have become a primary medium for presenting results, yet this valuable information remains 'locked' in a non-machine-readable image format. The ability to automatically extract and analyze data from these visualizations is therefore a fundamental step towards the comprehensive, large-scale analysis of scholarly literature.

While Large Language Models (LLMs) have emerged as a promising technology to address this challenge, their application is often hindered by reliability issues, such as the widely documented problem of factual 'hallucinations'. The non-deterministic, 'black box' nature of these models means that a single, complex prompt applied to a non-standardized input, such as a scientific chart, can lead to inconsistent or erroneous results [2]. This leads to a central research question: what is the most effective strategy to "help" an LLM read a chart more accurately?

Two primary classes of strategies can be considered. The first is semantic priming, which involves providing the model with high-level, contextual information to help it "understand" the chart's structure. This includes techniques such as Chain-of-Thought (CoT) prompting [3] or multi-stage workflows that first generate textual metadata about the chart. The second strategy is spatial priming, which involves augmenting the image with explicit visual cues to aid the model's low-level spatial localization capabilities, for instance, by overlaying a coordinate grid.

This paper presents a comparative investigation into the effectiveness of these two distinct strategies. Our findings reveal a counter-intuitive result: we demonstrate that semantic priming methods, including a two-stage "metadata-first" approach, failed to produce a statistically significant improvement over a simple baseline. In contrast, we show that the spatial priming method of applying a grid overlay provides a statistically significant reduction in data extraction error (SMAPE reduced from 25.5% to 19.5%, $p < 0.05$).

The primary contributions of this paper are therefore:

*1)* A comparative analysis of semantic versus spatial priming strategies for LLM-based chart data extraction.

*2)* Empirical evidence demonstrating that a simple grid overlay is a more effective method for improving accuracy than more complex, metadata-driven approaches for the current generation of models.

*3)* A discussion on the implications of these findings, suggesting that enhancing the model's low-level spatial perception is currently a more effective path than high-level semantic guidance for this specific task.

## II. Related Work

The task of automatically extracting data from charts has been approached from two main directions: classical computer vision techniques and, more recently, Large Language Model-based methods. Before presenting our proposed solution, it is crucial to position our work within this context, establish a performance baseline, and document the exploratory research that led to our final hypothesis.

### A. Related Work in Automated Chart Analysis

Prior to the LLM era, the problem was primarily addressed with classical computer vision (CV) and image processing algorithms. These methods often involve complex, multi-stage pipelines for detecting graphical elements, recognizing text, and reconstructing data, as exemplified by foundational work in reverse-engineering visualizations [4]. While powerful, such systems often rely on carefully engineered heuristics and can struggle with the high variability of charts found in real-world publications.

More recently, the advent of multimodal LLMs has led to the development of specialized models for chart understanding, such as Google's DePlot [5] and Microsoft's Chart-LLaMA [6]. These systems demonstrate impressive capabilities in end-to-end plot-to-table translation. However, their performance on non-standard charts remains an area of active research, and they often represent complex, monolithic architectures. In this paper, instead of focusing on model architecture, we investigate a different question: how can pre-processing strategies improve the performance of a general-purpose multimodal LLM on this task?

## III. Exploratory Analysis and Hypothesis Formulation

### A. The Baseline Approach

Our baseline serves as the control against which all other methods are measured. It represents a standard, straightforward implementation of a single-shot prompting strategy. In this approach, the original, unprocessed chart image is passed directly to the multimodal LLM (Gemini Pro) along with a carefully optimized prompt. The prompt instructs the model to perform all necessary cognitive tasks in a single pass. This baseline represents the standard performance level of the LLM without any external assistance.

### B. Exploratory Analysis: The Failure of High-Level Priming Strategies

Our initial investigation focused on two prominent hypotheses for improving upon the baseline by providing the model with high-level contextual information.

*1) Classical Computer Vision Pre-processing:* Our first attempt was to build a pre-processing pipeline inspired by established reverse-engineering techniques, using traditional CV algorithms (e.g., edge detection, contour analysis, and OCR) to programmatically extract the chart's metadata [7], [8]. The hypothesis was that providing the LLM with structured metadata would improve its accuracy. However, this approach failed on our dataset of real-world scientific charts, primarily due to their lack of standardization. Figure 1 provides a representative example of the challenges encountered.

As demonstrated in Figure 1, the variability in chart layout poses significant challenges. Our CV pipeline consistently failed on three key aspects:

*a) Legend Identification:* The pipeline was designed to find a distinct, bordered legend box. It was unable to reliably identify the floating text annotations ('Desert', 'Equatorial', etc.) as the legend because they lack a standard container.

*b) Line Following*: The occlusion of data series at multiple intersection points (highlighted by red circles) frequently confused line-following algorithms. A simple algorithm would often switch from tracking one data series to another after an intersection, leading to corrupted coordinate data.

*c) Label-to-Data Association:* Even if the text and lines were detected separately, semantically associating each floating label with its correct data series requires complex spatial heuristics (e.g., "find the nearest line") that are highly prone to error.

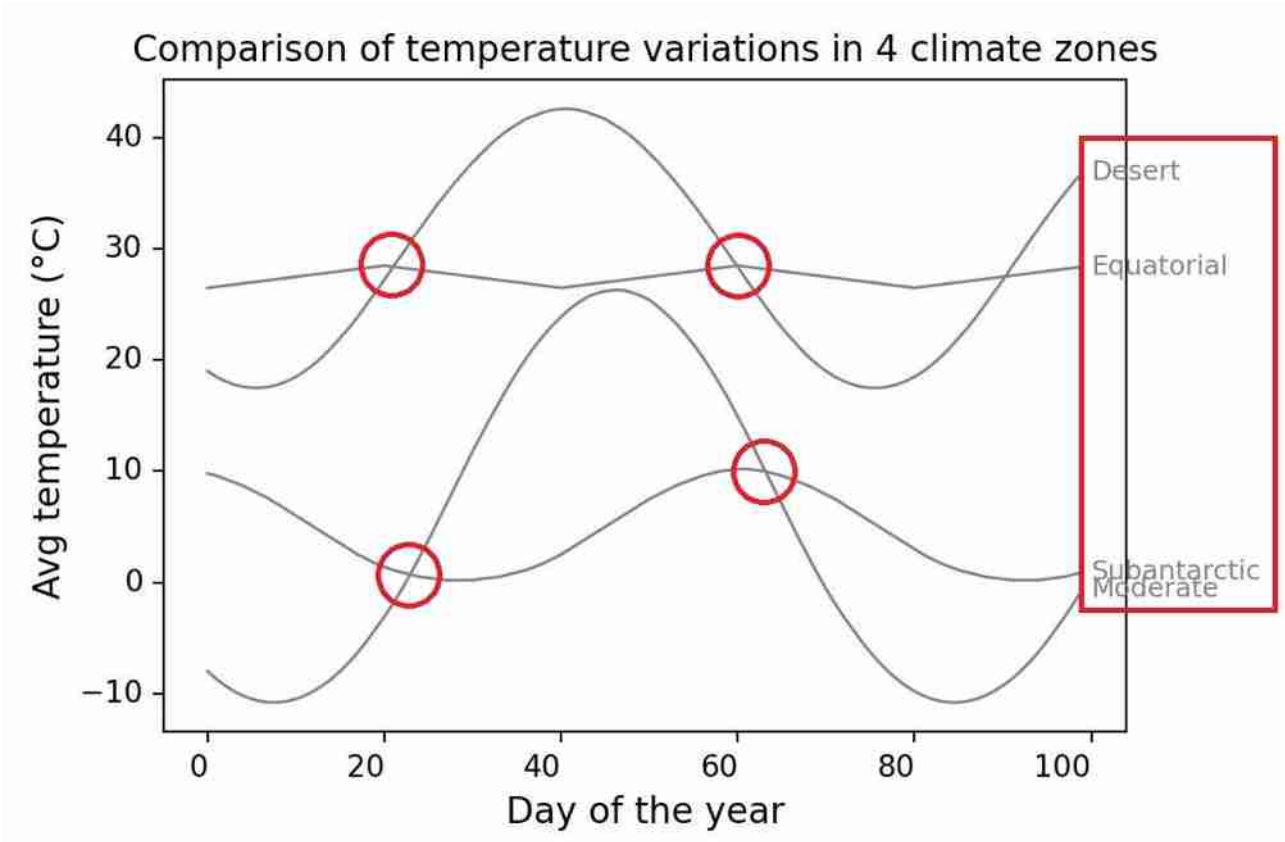

Fig. 1. A synthetic chart illustrating key failure modes for classical CV algorithms. The lack of a formal legend box prevents reliable label identification. The use of floating text annotations complicates label-to-data association. Finally, the occlusion of data series at intersection points (circled) leads to corrupted line following.

Providing this flawed, algorithmically generated metadata to the LLM was found to be more detrimental than providing no metadata at all, as it actively misled the model. This initial finding motivated our decision to use a controlled synthetic dataset for the final quantitative experiment to precisely isolate variables.

*2) Semantic Priming via LLM:* Our second hypothesis was that the LLM itself could generate more reliable semantic context. We explored two methods: a two-stage "metadata-first" framework and single-shot Chain-of-Thought (CoT) prompting [3]. In our preliminary tests, however, **neither of these semantic approaches yielded a statistically significant improvement** over the baseline.

The systematic failure of methods aimed at high-level "understanding" (both classical and LLM-based) led us to an alternative hypothesis: the primary limitation for this task may not be high-level reasoning, but rather low-level spatial localization. This insight prompted the development of the novel spatial priming framework, which is detailed in the following section.

## IV. The Proposed Grid-Based Framework

Based on the findings that high-level priming strategies are ineffective, we developed a novel framework grounded in our alternative hypothesis: that performance is primarily limited by low-level spatial localization. This section details the rationale and architecture of this grid-based approach, which is designed to provide the LLM with explicit visual cues.

## A. Rationale and Hypothesis

The core rationale behind our grid-based approach is grounded in a foundational principle in computer vision: that dividing a complex image into a grid of smaller regions is an effective strategy for extracting robust local features [9]. While classical algorithms used this for feature engineering, we adapt this principle as a pre-processing step for modern LLMs.

Our approach can also be viewed as providing a form of **explicit spatial priming**, which is analogous to the internal mechanisms of state-of-the-art models. Vision Transformers (ViT), for instance, operate by partitioning an image into a grid of patches to which they apply positional encodings [10]. Our external grid overlay serves a similar purpose: it provides the model with a clear, unambiguous coordinate system across the entire image.

This transforms the task of data extraction from an implicit estimation problem into an explicit coordinate identification task. Without a grid, the LLM must infer a point's location relative to sparsely labeled tick marks. With a grid, it is tasked with a simpler problem: identifying which grid cell a point falls into.

Our central hypothesis is that providing this explicit spatial context reduces the model's cognitive load and mitigates ambiguity in spatial localization. We posit that this will lead to a statistically significant reduction in data extraction error.

## B. Implementation and Architecture

The implementation of the framework is intentionally straightforward, consisting of a single pre-processing step applied before the standard prompting procedure.

The method is as follows:

*1) Grid Overlay:* Before the chart image is passed to the LLM, we programmatically overlay a semi-transparent 50x50 coordinate grid onto it using a standard image processing library. The grid consists of 50 vertical and 50 horizontal lines, dividing the image into 2500 individual cells. The lines are rendered with a low opacity (e.g., 20%) to ensure they do not obscure the underlying data series.

*2) Standard Prompting:* This modified image, now containing the grid, is then passed to the multimodal LLM (Gemini Pro) using the identical single-shot prompt as the baseline system. No changes are made to the prompt itself.

The architecture of this process is illustrated in Figure 2.

In essence, our proposed method does not alter the LLM's reasoning process via complex instructions but instead modifies the input data itself to make the spatial localization task easier for the model. The following section details the experimental setup designed to quantitatively measure the impact of this spatial priming approach on data extraction accuracy.

# V. Experimental Setup

This section details the experimental design used to evaluate our proposed framework. We describe the dataset and the process of creating the gold standard, introduce the systems used for comparison, and define the evaluation metrics.

## A. Source Dataset

While real-world charts motivated this study, a synthetic dataset was chosen for the quantitative experiment to ensure a controlled and perfectly accurate Gold Standard, allowing for a precise measurement of the grid overlay's impact.

Our evaluation is conducted on a purpose-built dataset comprising 23 diverse line graphs. This dataset was synthetically generated from source JSON data, with each data series containing 100 data points. The design of these graphs is intended to replicate the most common types of line graphs found in scientific articles and journals.

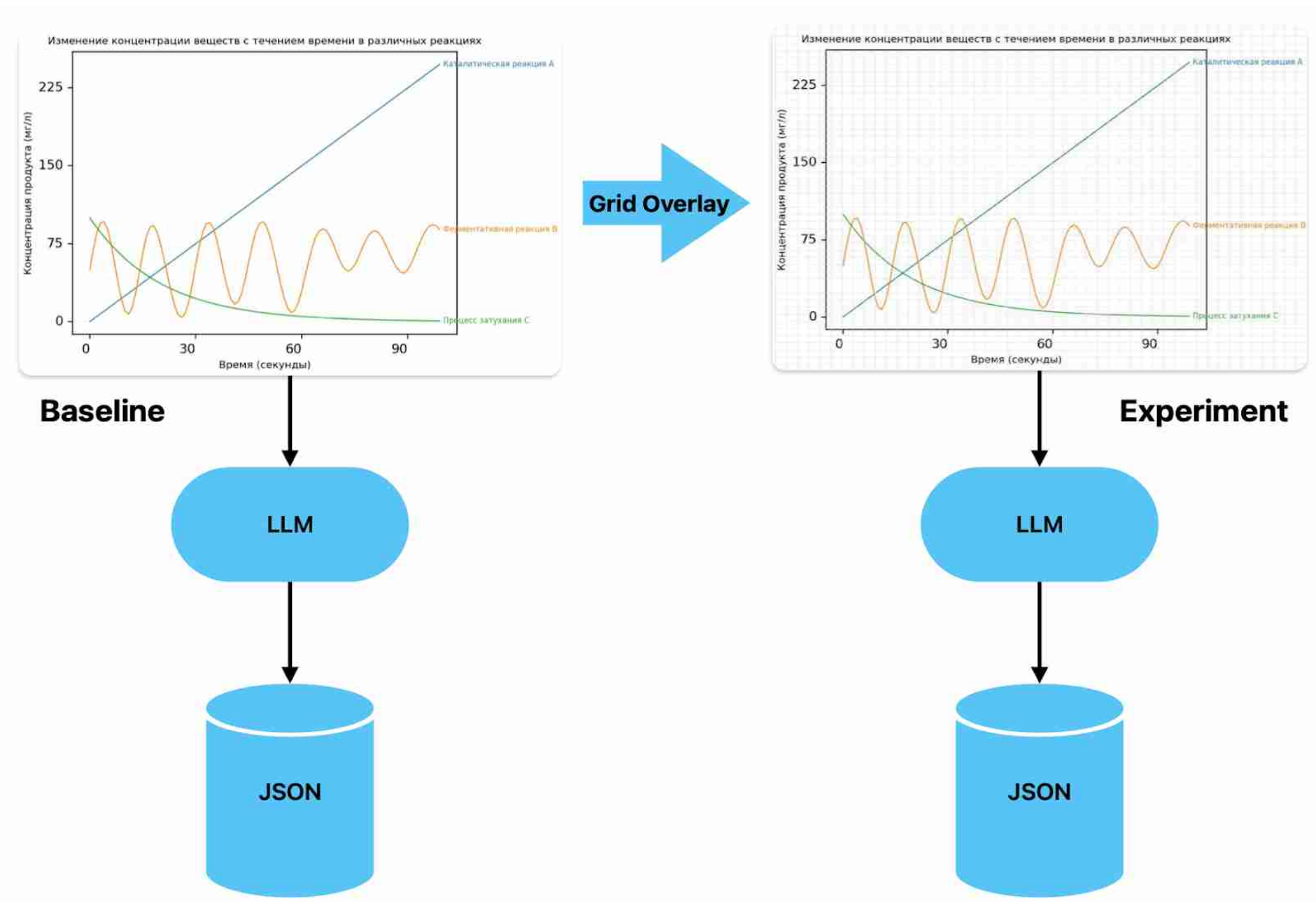

Fig. 2. The architecture of the proposed spatial priming framework, where a pre-processing step applies a grid overlay before the image is passed to the LLM.

To ensure comprehensive evaluation, the dataset includes a variety of styles, such as:

- Black and white graphs.
- Graphs with legends embedded within the plot area.
- Graphs with external legends.
- Plots with and without gridlines.

This diversity allows for a robust assessment of data extraction performance across a range of visually distinct scenarios. The examples of source graphs are illustrated on Figure 3.

To establish a reliable ground truth for our experiments, a Gold Standard was created not by manually extracting the data points from each of the 23 graphs, but from their source JSON data with their 100 points per data series used to generate each figure. This method guarantees that Gold Standard data is 100% accurate and mitigates human error.

## B. *Systems for Comparison*

To evaluate the effectiveness of our proposed method, we compare it against a baseline approach.

- **Baseline Approach**: This is a single-step method that utilizes a direct prompt to instruct the model to extract all coordinates and relevant metadata from the graph image. The model is expected to return a JSON object containing the chart's label, axis information, and a list of data series, where each series includes its name and an array of (X,Y) coordinate pairs.
- **Our Proposed Framework**: This method utilizes the spatial priming approach detailed in Section III. It consists of a pre-processing step where a 50x50 grid is overlaid onto the chart image. This modified image is then passed to the LLM using the same extraction prompt as the baseline. The output is also a JSON object with the same structure.

## C. *Evaluation Methodology*

To provide a robust quantitative comparison between the baseline and our proposed framework, we established a formal evaluation methodology.

The primary metric for accuracy is the **Symmetric Mean Absolute Percentage Error (SMAPE)**. This metric was chosen over standard MAPE as it provides a more balanced measure of error on data that may contain zero or near-zero values. The formula (1) is as follows:

$$SMAPE = \frac{100\,\%}{n} \sum \frac{|F - A|}{(|A| + |F|)/2} \quad (1)$$

where n is the number of data points, F (Forecast) is the extracted value, and A (Actual) is the value from the Gold Standard.

To enable a direct, point-by-point comparison between the sparse coordinate set generated by the LLM and the dense 100-point Gold Standard, we first apply linear interpolation to the LLM's output. This process generates a curve with the same 100 X-coordinates as the Gold Standard, allowing for a comprehensive calculation of the SMAPE across the entire dataset.

After calculating the SMAPE score for each of the 23 graphs for both methods, we employ the Wilcoxon signed-rank test to determine if the observed performance difference is statistically significant. This non-parametric test was chosen over a paired t-test because it does not assume that the differences between the paired SMAPE scores follow a normal distribution, making it more suitable for analyzing error metrics. We reject the null hypothesis (that the median difference between the methods is zero) if the calculated p-value is below a significance level of $\alpha = 0.05$.

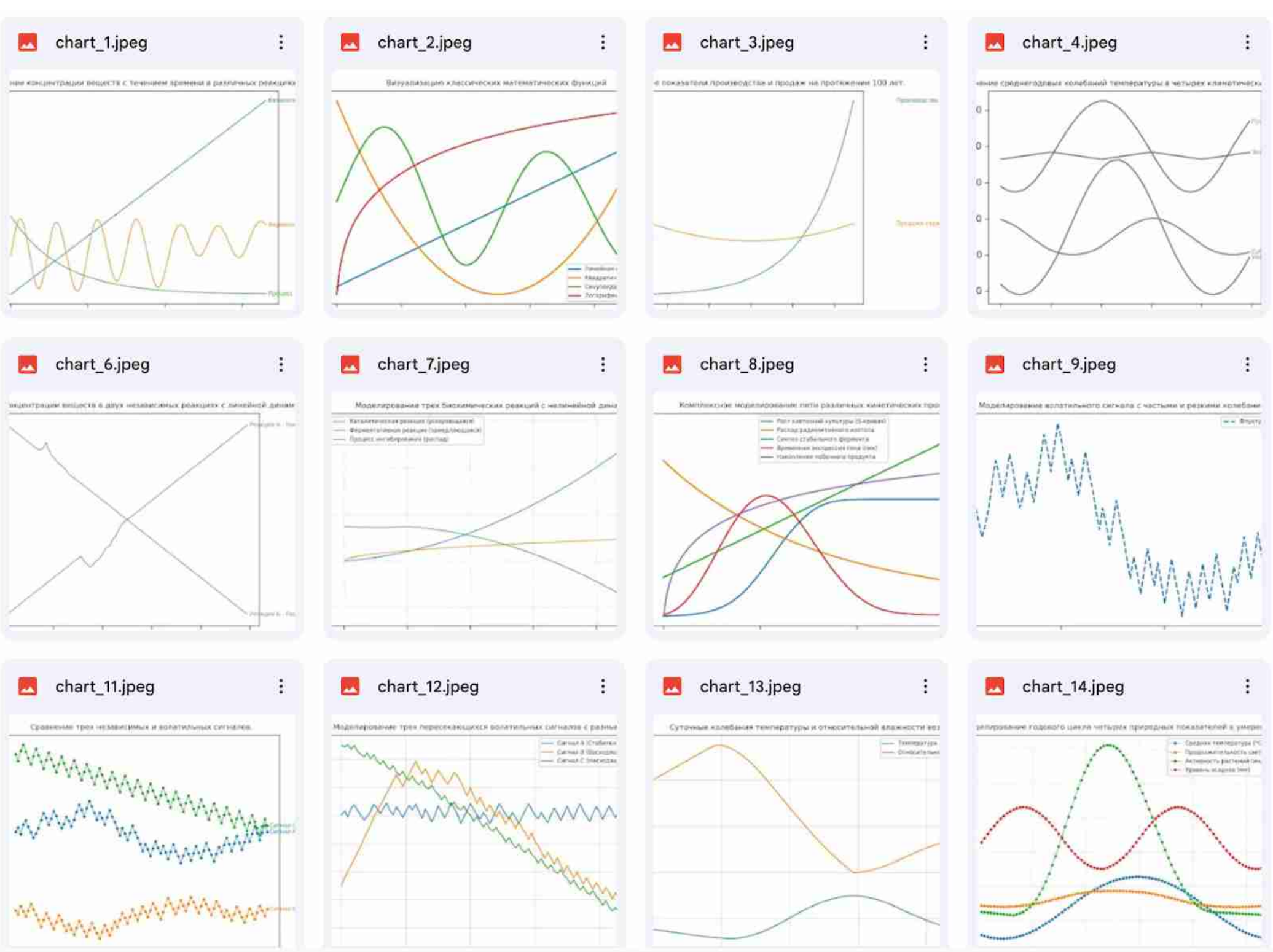

Fig. 3. Examples of source graphs

## VI. Results and Discussion

This section presents the quantitative results of our comparative experiment, followed by a discussion of the key findings and their implications.

### A. *Quantitative Results*

The performance of our proposed grid-based framework was compared against the baseline approach using the Symmetric Mean Absolute Percentage Error (SMAPE) metric, where a lower value indicates higher accuracy. The aggregated results, including the mean and standard deviation of the SMAPE scores across our dataset, are summarized in Table 1.

TABLE I. Comparative Accuracy and Reliability of Data Extraction

| Method | Mean SMAPE,% | Std. Dev., % |
|---|---|---|
| Baseline | 25.48 | 26.01 |
| Experimental (Grid) | 19.48 | 14.61 |

As shown in Table 1 and clearly illustrated in Figure 4, our proposed method achieved a mean SMAPE of 19.48%, representing a substantial improvement over the Baseline approach (mean SMAPE of 25.48%). A Wilcoxon signed-rank test confirmed that this difference is **statistically significant (p = 0.03)**.

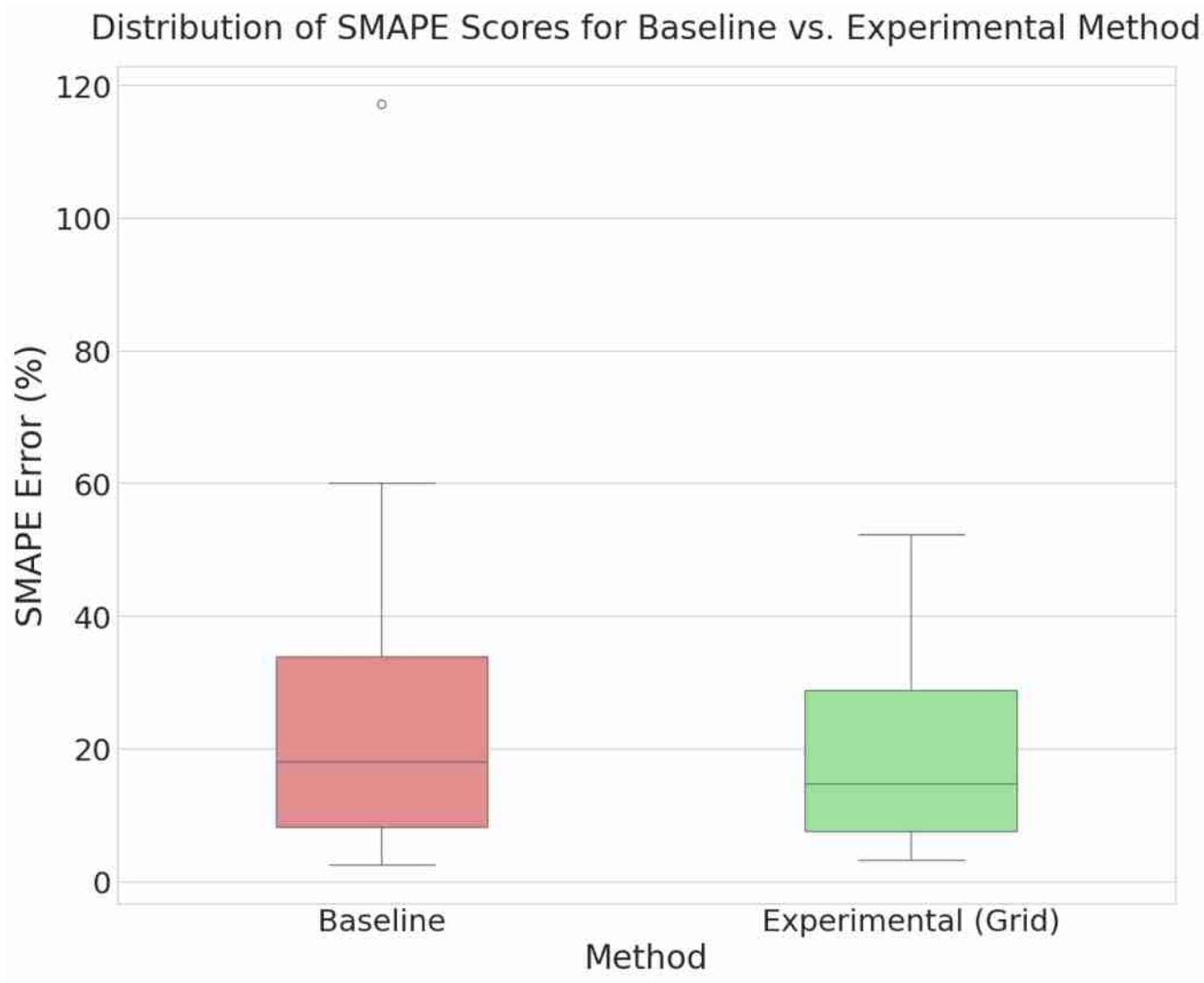

Fig. 4. Distribution of SMAPE Scores for Baseline vs. Experimental Method. The plot shows the median (center line), interquartile range (box), and overall range (whiskers) of the error scores. The outlier for the Baseline is also shown.

Furthermore, the analysis of the data distribution reveals a critical difference in reliability. The Baseline approach exhibited extremely high variance (Std. Dev. = 26.01), a fact visually represented in Figure 4 by its large interquartile range, long whisker, and a significant outlier. This indicates highly inconsistent performance. In contrast, our Experimental (Grid) method shows a significantly lower standard deviation (14.61) and a much more compact distribution in the box plot, demonstrating a more stable and predictable performance.

Therefore, we conclude that the grid overlay not only improves the average accuracy but, more critically, **enhances the method's reliability** by mitigating the risk of severe errors.

### B. *Discussion*

While the aggregated metrics in Table 1 and the distributional analysis in Figure 4 demonstrate the superior reliability of our method, a qualitative analysis of a representative sample provides a clearer picture of the performance difference in terms of accuracy. Figure 5 illustrates a direct comparison of the two methods on a single, complex chart from our dataset.

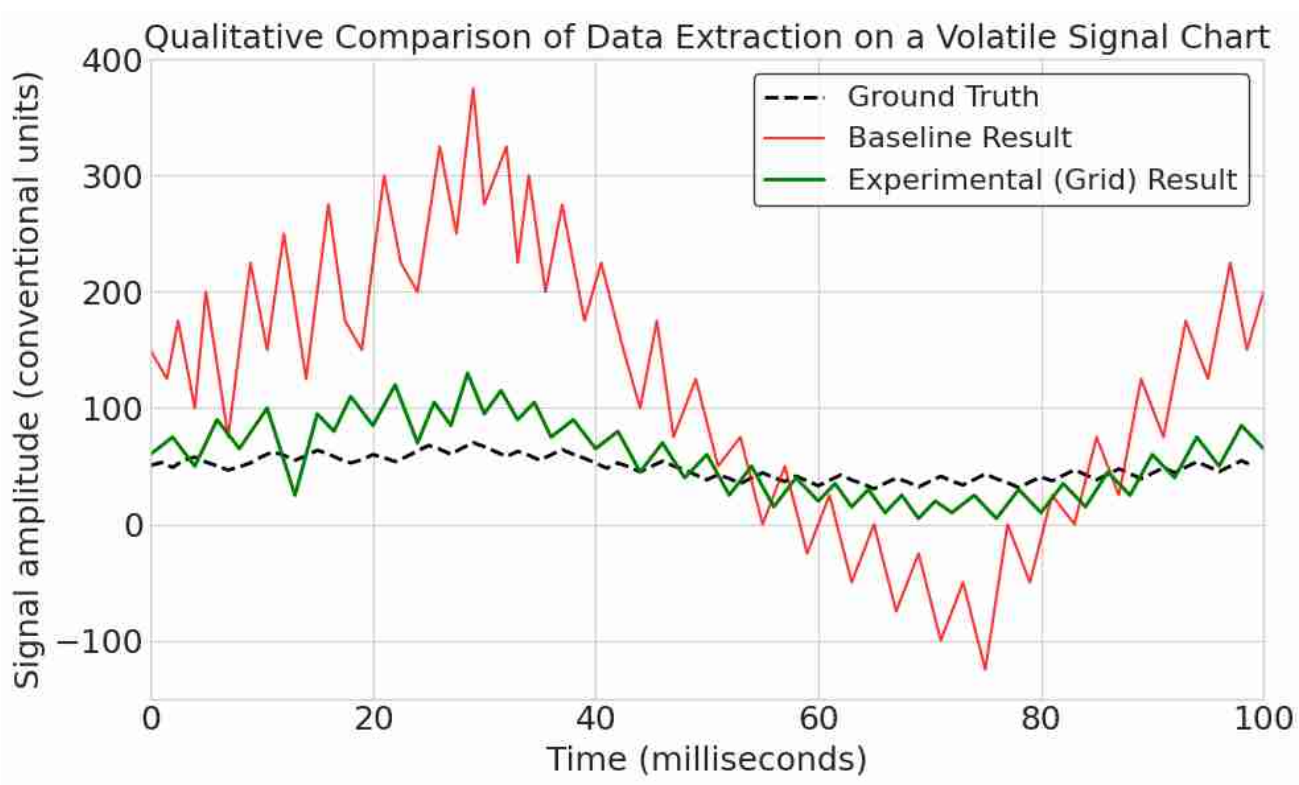

Fig. 5. Qualitative Comparison of Data Extraction Results on a Volatile Signal Chart. The plot compares the Ground Truth data (black dashed line) against the interpolated curves generated from the outputs of the Baseline method (red line) and our proposed Experimental (Grid) method (green line).

Figure 5 demonstrates that the Baseline result (red line) fails to capture the nuanced dynamics of the data. It exhibits excessive volatility, inventing sharp peaks and troughs that are not present in the source data, and significantly deviates from the ground truth, particularly in the 70-80ms range. In stark contrast, our grid-based method (green line) closely tracks the Ground Truth curve across its entire length. This visual evidence highlights that the improvement provided by our method is not merely numerical, but qualitative: our method produces a more faithful and reliable representation of the source data.

This clear qualitative difference in performance warrants a deeper discussion of the underlying reasons for the grid method's success and the semantic methods' failure.

Our results suggest two primary interpretations of the model's behavior.

*1) The Effectiveness of Spatial Priming:* The success of the grid overlay supports our hypothesis that the primary bottleneck for this task is low-level spatial localization. This approach is grounded in foundational computer vision principles, where dividing an image into a grid is a proven strategy for extracting robust local features [9]. By making this grid explicit, we provide the LLM with a clear coordinate system, a process analogous to the internal patch-based mechanisms of modern Vision Transformers [10]. This transforms an implicit estimation problem ("Where is this point?") into a more concrete identification task ("In which grid cell is this point?"), thereby "grounding" the model's spatial reasoning and improving precision.

*2) The Ineffectiveness of Semantic Cues:* Perhaps the most interesting finding is that improvements in low-level spatial localization (via the grid) yielded better results than high-level semantic understanding (metadata generation or CoT). Our results suggest that current multimodal models may benefit more from enhancements that aid in precise visual tasks rather than from rich semantic prompts. It is likely that the internal mechanisms for integrating visual analysis with semantic reasoning are not yet sufficiently advanced. In this context, complex semantic cues may act as a distraction, potentially interfering with the model's ability to perform a direct and accurate visual analysis of the graph's data.

## VII. Limitations and Future Work

Although our findings provide strong evidence for the effectiveness of the grid-based approach, we acknowledge several limitations that define the scope of this study and offer clear directions for future research.

### A. Limitations

*1) Dataset Composition:* Our experiment was conducted on a synthetically generated dataset. While this approach was deliberately chosen to ensure a perfectly accurate Gold Standard for precise error measurement, it does not capture the full spectrum of noise and variability present in "in-the-wild" scientific charts (e.g., compression artifacts, hand-drawn annotations, scanning imperfections). Therefore, the performance uplift we observed should be considered an upper bound, and further validation on a dataset of real-world charts is required.

*2) Image Size Constraints:* Our preliminary tests indicated that multimodal LLMs internally downscale large images, potentially leading to a loss of fine-grained details. To mitigate this, our study was limited to images with a maximum dimension of 1200 pixels. This means our findings may not be directly applicable to very high-resolution charts, and the interaction between image resolution, grid density, and model performance remains an open question.

*3) 3. Fixed Grid Density:* This study utilized a fixed 50x50 grid for all experiments. While effective, this density was chosen heuristically. It is likely not the optimal configuration for all types of charts; for instance, charts with very dense data may require a finer grid, while simpler charts might perform better with a coarser one.

### B. Future Work

Building on the insights and limitations of this study, our future work will proceed along three key tracks:

*1) Optimization of Grid Parameters:* The most immediate next step is to conduct a systematic analysis of the grid's hyperparameters. This includes investigating the optimal grid density (e.g., comparing 25x25, 50x50, and 100x100 grids) and the effect of line opacity and color on model performance. The goal is to develop a set of best practices or even an adaptive method for grid generation.

*2) Development of a Hybrid, Content-Aware Framework:* A more advanced approach would be to combine the strengths of classical CV and our spatial priming method. Future work will explore a hybrid framework where a simple CV algorithm first detects the precise boundaries of the plot area. The grid would then be applied only to this area, avoiding the creation of visual noise over the title, legend, and axis labels. This content-aware approach could further improve accuracy.

*3) Validation on Real-World Data at Scale:* To validate the real-world applicability of our findings, we plan to assemble a large-scale, annotated dataset of charts from diverse scientific publications. Evaluating our grid-based method on this dataset will be the ultimate test of its robustness and generalizability.

## VIII. Conclusion

This paper presented a rigorous investigation into strategies for improving the accuracy of data extraction from scientific charts using Large Language Models. We systematically evaluated two distinct classes of approaches: high-level semantic priming (including metadata-generation and Chain-of-Thought) and low-level spatial priming (via a grid overlay).

Our central contribution is the empirical demonstration of a counter-intuitive finding: a simple spatial intervention significantly outperforms more complex semantic methods. Our experiments showed that overlaying a coordinate grid onto the chart image yielded a statistically significant reduction in error (SMAPE reduced from 25.5% to 19.5%, $p < 0.05$), whereas the various semantic priming techniques failed to produce a significant improvement over the baseline.

The key implication of this work is that for the current generation of multimodal models, enhancing low-level spatial perception is a more effective and reliable strategy than high-level semantic guidance for this class of tasks. This suggests that the primary bottleneck is not the model's ability to "understand" a chart, but its ability to precisely "see" it. Ultimately, this paper provides a validated, practical methodology for engineers and researchers, suggesting that the path to more robust AI-driven analysis may lie less in the complexity of the prompt and more in the intelligent pre-processing of the input data itself.